%% file: template.tex
\RequirePackage{fix-cm}
\documentclass[twocolumn]{svjour3}          
\smartqed  
\usepackage{amsmath}
\usepackage{graphicx}
\usepackage{textcomp}
\usepackage{hyperref}
\usepackage[USenglish]{babel}
\usepackage[TABTOPCAP]{subfigure}
\usepackage[utf8]{inputenc}
\usepackage{color}
\usepackage{ifthen}                                                            
\newboolean{showNotes}  
\setboolean{showNotes}{false}
\newcommand{\todo}[1]{\ifthenelse{\boolean{showNotes}}{\textcolor{red}{\textbf{\textcolor{red}{(TODO: #1)}}}}{}}

\newboolean{showConvnet1Curvefig}
\setboolean{showConvnet1Curvefig}{false}
\newboolean{showFilters}
\setboolean{showFilters}{true}

\newcommand{\cut}[1]{}

\newcommand{\comment}[1]{}

\def\best14{47.67}

\begin{document}
\title{EmoNets: Multimodal deep learning approaches for emotion recognition in video}

\author{Samira Ebrahimi Kahou
\and
        Xavier Bouthillier
\and
        Pascal Lamblin
\and
        Caglar Gulcehre
\and 
        Vincent Michalski
\and
        Kishore Konda
\and
        Sébastien Jean
\and
        Pierre Froumenty
\and
        Yann Dauphin
\and 
        Nicolas Boulanger-Lewandowski
\and 
        Raul Chandias Ferrari
\and 
        Mehdi Mirza 
\and
        David Warde-Farley
\and
        Aaron Courville 
\and
        Pascal Vincent 
\and
        Roland Memisevic 
\and
        Christopher Pal 
\and
        Yoshua Bengio 
}


\institute{\begin{flushleft}S.E. Kahou, P. Froumenty, C. Pal \at
              \'Ecole Polytechique de Montr\'eal, Universit\'e de Montr\'eal, Montr\'eal, Canada
              \at
              Email: \{samira.ebrahimi-kahou, pierre.froumenty, christopher.pal\}@polymtl.ca \end{flushleft}
\begin{flushleft}
           V. Michalski, K. Konda \at
           Goethe-Universität Frankfurt, Frankfurt, Germany
           \at
           Email: \{michalskivince, konda.kishorereddy\}@gmail.com \end{flushleft}
           \begin{flushleft}
           X. Bouthillier, P. Lamblin, C. Gulcehre, S. Jean,
           Y. Dauphin, N. Boulanger-Lewandowski,
           R.C. Ferrari, M. Mirza, D. Warde-Farley,
           A. Courville, P. Vincent, R. Memisevic, Y. Bengio \at
              Laboratoire d'Informatique des Syst\`emes Adaptatifs, Universit\'e de Montr\'eal, Montr\'eal, Canada 
              \at
              Email: \{bouthilx, lamblinp, gulcehrc, jeasebas, dauphiya, boulanni, chandiar, mirzamom, wardefar,
              courvila, vincentp, memisevr,  bengioy\}@iro.umontreal.ca \end{flushleft}
}

\maketitle

\begin{abstract}
The task of the emotion recognition in the wild (EmotiW) Challenge is to assign one of seven emotions to short video clips extracted from Hollywood style movies. The videos depict acted-out emotions under realistic conditions with a large degree of variation in attributes such as pose and illumination, making it worthwhile to explore approaches which consider combinations of features from multiple modalities for label assignment.

In this paper we present our approach to learning several specialist models using deep learning techniques, each focusing on one modality. Among these are a convolutional neural network, focusing on capturing visual information in detected faces, a deep belief net focusing on the representation of the audio stream, a K-Means based ``bag-of-mouths'' model, which extracts visual features around the mouth region and a relational autoencoder, which addresses spatio-temporal aspects of videos.

We explore multiple methods for the combination of cues from these modalities into one common classifier. This achieves a considerably greater accuracy than predictions from our strongest single-modality classifier. Our method was the winning submission in the 2013 EmotiW challenge and achieved a test set accuracy of \best14\% on the 2014 dataset.


\end{abstract}

\section{Introduction}
\label{intro}
This is an extended version of the paper describing our winning submission \cite{kahou2013combining} to the Emotion Recognition in the Wild Challenge (EmotiW) in 2013 \cite{EmotiW}. Here we describe our approach in more detail and present results on the new data set from the 2014 competition  \cite{dhall2014emotion}.
The task in this competition is to assign one of seven emotion labels (angry, disgust, fear, happy, neutral, sad, surprise) to each short video clip in the Acted Facial
Expression in the Wild (AFEW) dataset \cite{dhall2012collecting}. 
The video clips are extracted from feature films. Given the low number of samples per emotion category, it is difficult to deal with the large variety of subjects, lighting conditions and poses in these close-to-real-world videos.
The clips are approximately 1 to 2 seconds long and also feature an audio track, which might contain voices and background music. 

We explore different methods of combining predictions of modality-specific models, including: (1) a deep convolutional neural network (ConvNet) trained to recognize facial expressions in single frames; (2) a deep belief net that is trained on audio information; (3) a relational autoencoder that learns spatio-temporal features, which help to capture human actions; and (4) a shallow network that is trained on visual features extracted around the mouth of the primary human subject in the video. 
We discuss each model, their performance characteristics and different aggregation strategies. The best single model, without considering combinations with other experts, is the ConvNet trained to predict emotions given still frames. It has been trained only on additional facial expression datasets, i.e. not using the competition data. The ConvNet was then used to extract class probabilities for the competition data. 
The extracted  probability vectors of the challenge training and validation sets were aggregated to fixed-length vectors and then used to train and validate hyperparameters of a support vector machine (SVM) for final classification. This yielded a test set accuracy of 35.58\% for the 2013 dataset.
Using our best strategy (at the time) for the combination of top performing expert models into a single predictor, we were able to achieve an accuracy of 41.03\% on the 2013 challenge test set. The next best competitor achieved a test accuracy of 35.89\%.
We reran our pipeline on the 2014 challenge data with improved settings for our combination model and achieved a test set accuracy of \best14\%, compared to 50.37\% reported by the challenge winners \cite{liu2014combining}.

\section{Related work}
\label{relatedwork}
The task of recognizing the emotion to associate with a short video clip is well suited for methods and models that combine features from different modalities. As such, many other successful approaches in the Emotion recognition in the Wild (EmotiW) 2013 and 2014 challenges focus on the fusion of modalities. These include \cite{sikka2013multiple}, who used Multiple Kernel Learning (MKL) for fusion of visual and audio features. 
The recent success of deep learning methods in challenging computer vision \cite{NIPS2012_0534}\cut{\cite{szegedy2014going}}\cite{neverova2014moddrop}\cite{kahoufacial}, language modeling \cite{kalchbrenner2014convolutional} and speech recognition \cite{hinton2012deep} tasks seems to carry over to emotion recognition, taking into account that the 2014 challenge winners \cite{liu2014combining} also employed a deep convolutional neural net, which they combined with other visual and audio features using a Partial Least Squares (PLS) classifier. 
The adoption of deep learning for visual features likely played a big role in the considerable improvement compared to their submission in the 2013 competition \cite{liu2013partial}, although the first and second runners up also reached quite good performances without deep learning methods; \cite{sun2014combining} used a hierarchical classifier for combining audio and video features and \cite{chen2014emotion} introduced an extension of Histogram of Oriented Gradients (HOG) descriptors for spatio-temporal data, which they fuse with other visual and audio features using MKL.

\section{Models for modality-specific representation learning}
\label{sec:models}
\subsection{A convolutional network approach for faces}
\label{sec:models-conv1}
\input{convnet1}

\input{audio.tex}

\subsection{Activity recognition using a relational autoencoder}
\label{sec:models-activity}
\input{activity.tex}
\subsection{Bag of mouth features and shallow networks}
\label{sec:models-bom}

\input{bom.tex}

\section{Experimental results}
\label{sec:combine}
\input{combination.tex}

\begin{table}
\caption{Test accuracies of different approaches on AFEW2 (left) and AFEW4 (right)}

\label{tab:afew2}
\begin{tabular}{ll}
\hline\noalign{\smallskip}
Method & \%  \\
\noalign{\smallskip}\hline\noalign{\smallskip}
 MKL \cite{sikka2013multiple} & 35.89\% \\
 PLS \cite{liu2013partial} & 34.61\% \\
 Linear SVM \cite{gehrig2013facial} &  29.81\% \\
\noalign{\smallskip}\hline
 Our method \cite{kahou2013combining} & 41.03\% \\
\noalign{\smallskip}\hline
\end{tabular}
\quad
\begin{tabular}{ll}
\hline\noalign{\smallskip}
Method & \%  \\
\noalign{\smallskip}\hline\noalign{\smallskip}
  PLS \cite{liu2014combining} & 50.37\% \\
  HCF \cite{sun2014combining} & 47.17\% \\
  MKL \cite{chen2014emotion} & 45.21\% \\
\noalign{\smallskip}\hline
 Our method & \best14\% \\
\noalign{\smallskip}\hline
\end{tabular}
\vspace{-.3cm}
\end{table}

\section{Conclusions and discussion}
\input{conclusion.tex}

\noindent\textbf{Acknowledgements}
The authors would like to thank the developers of Theano \cite{bastien2012theano,bergstra2010theano}.
We thank NSERC, Ubisoft, the German BMBF, project 01GQ0841 and CIFAR for their support. 
We also thank 
Abhishek Aggarwal,
Emmanuel Bengio,
Jörg Bornschein,
Pierre-Luc Carrier,
Myriam Côté,
Guillaume Desjardins,
David Krueger,
Razvan Pascanu,
Jean-Philippe Raymond,
Arjun Sharma,
Atousa Torabi,
Zhenzhou Wu,
and
Jeremie Zumer 
for their work on 2013 submission.


\cut{
Abstract
Keywords
1. Introduction
2. Related work
3. Datasets [all extra data and AFEW]
4. Models   [what is convnet, RBM, Deep belief etc]
5. Results  [how we trained each model in detail with confusion matrices]
6. Model combination methods
7. Discussion
Acknowledgements
References

}

\bibliographystyle{spmpsci}
\bibliography{template}

\end{document}

%% file: convnet1.tex
ConvNets are artificial neural network architectures, that assume a topological 
input space, e.g. a 2d image plane. A set of two-dimensional or 
three-dimensional (if the inputs are color images) filters is applied to small 
regions over the whole image using convolution, yielding a bank of filter 
response maps (one map per filter), which also exhibit a similar 2d topology. 

To reduce the dimensionality of feature banks and to introduce invariance
with respect to slight translations of the input image, convolutional layers 
are often followed by a pooling layer, which subsample the feature maps by collapsing 
small regions into a single element (for instance by choosing the maximum or 
mean value in the region).
ConvNets have recently been shown to achieve state of the art performance in challenging object recognition tasks \cite{NIPS2012_0534}. 

Because of the small number of training samples, our initial experiments with ConvNets showed severe overfitting on the 
training set, achieving an accuracy of $96.73\%$ on the AFEW2 training set, compared to only $35.32\%$ on the validation set. 
For this reason we decided to train on a separate dataset, which we refer to as 'extra data'. It consists of two face image datasets and is described in Section \ref{sec:extradata}.

The approach for the face modality can roughly be divided into four stages:
\begin{enumerate}
  \item Training the ConvNet on faces from extra data. The architecture is described in Section \ref{sec:models-conv1-conv1}.
  \item Extraction of 7-class probabilities for each frame of the facetubes (described in
    Section \ref{sec:models-conv1-facetube}).
  \item Aggregation of single frame probabilities into fixed-length video descriptors for each video in the competition dataset by expansion or contraction.
  \item Classification of all video-clips using a support vector machine (SVM) trained on video descriptors of the competition training set.
\end{enumerate}

\begin{figure*}
  \centering
  \includegraphics[width=.9\textwidth]{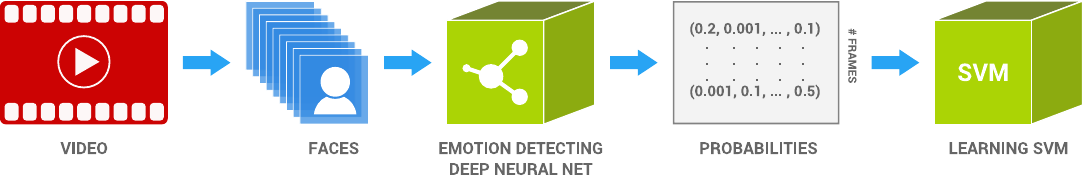}
\caption{Complete pipeline describing the final strategy used for our ConvNet \textnumero 1 model.}
\label{fig:pipeline_convnet}       
\end{figure*}

\noindent Stage three and four are described in detail in Section \ref{sec:models-conv1-svm}.
The pipeline is depicted in Figure \ref{fig:pipeline_convnet}.
The strategy of training on extra data and using the competition data only for classifier training and early stopping yielded a much lower training set accuracy of $46.87\%$, but it
achieved a considerably better validation set accuracy of $38.96\%$.

\subsubsection{Additional Face Dataset}
\label{sec:extradata}
The 'extra data' we used for training of the deep network is composed of two 
large static image datasets of facial expressions for the seven emotion classes. 

The first and larger one is the Google dataset \cite{googledata} 
consisting of 35,887 images with the seven facial expression classes: 
angry, disgust, fear, happy, sad, surprise and neutral. 
The dataset was built by harvesting images returned from Google’s image search 
using keywords related to expressions, then cleaned and labeled by hand. 
We use the grayscale $48\times 48$ pixel versions of these images.
The second one is the Toronto Face Dataset (TFD) \cite{TFD} containing 
4,178 images labeled with basic emotions, essentially with only fully frontal 
facing poses. 

To make the datasets compatible (there are big differences, for instance 
variation among subjects, lighting and poses), we applied the following 
registration and illumination normalization strategies:

\paragraph{Registration} To build a common dataset, TFD images and frames
from the competition dataset had to be integrated with the Google dataset, for
which we used the following procedure:
For image registration we used $51$ of the $68$ facial keypoints extracted by 
the mixture of trees method from \cite{ZhuRamFace}. The face contour keypoints 
returned by this model were ignored in the registration process. Images from 
the Google dataset and the AFEW datasets have different poses, but most faces are frontal 
views. 

To reduce noise, the mean shape of frontal pose faces for each dataset was used to 
compute the transformation between the two shapes. For the transformation the 
Google data was considered as base shape and the similarity transformation was used 
to define the mapping. After inferring this mapping, all data was mapped to the
Google data. TFD images have a tighter fit around faces, while Google data 
includes a small border around the faces. To make the two datasets compatible, 
we added a small noisy border to all images of TFD.

\paragraph{Illumination normalization using isotropic smoothing}

\begin{figure}
  \centering
  \includegraphics[width=.98\columnwidth]{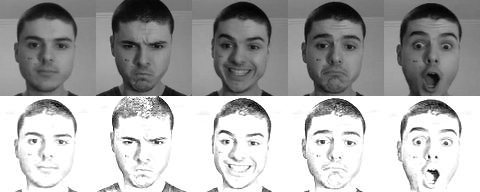}
\caption{Raw images at the top and the corresponding IS-preprocessed images below.}
\label{fig:is_smoothing}
\end{figure}

To compensate for varying illumination in the merged data\-set, we used the
diffusion-based approach introduced in \cite{heusch2005lighting}. We used the isotropic 
smoothing (IS) function from the INface toolbox \cite{vstruc2009gabor,IGI2011} 
with the default smoothness parameter and without normalization as 
post-processing. A comparison of original and IS-pre\-pro\-cessed face images is shown 
in figure \ref{fig:is_smoothing}.

\subsubsection{Extracting frame-wise emotion probabilites}
\label{sec:models-conv1-conv1}

\begin{figure*}
  \includegraphics[width=.98\textwidth]{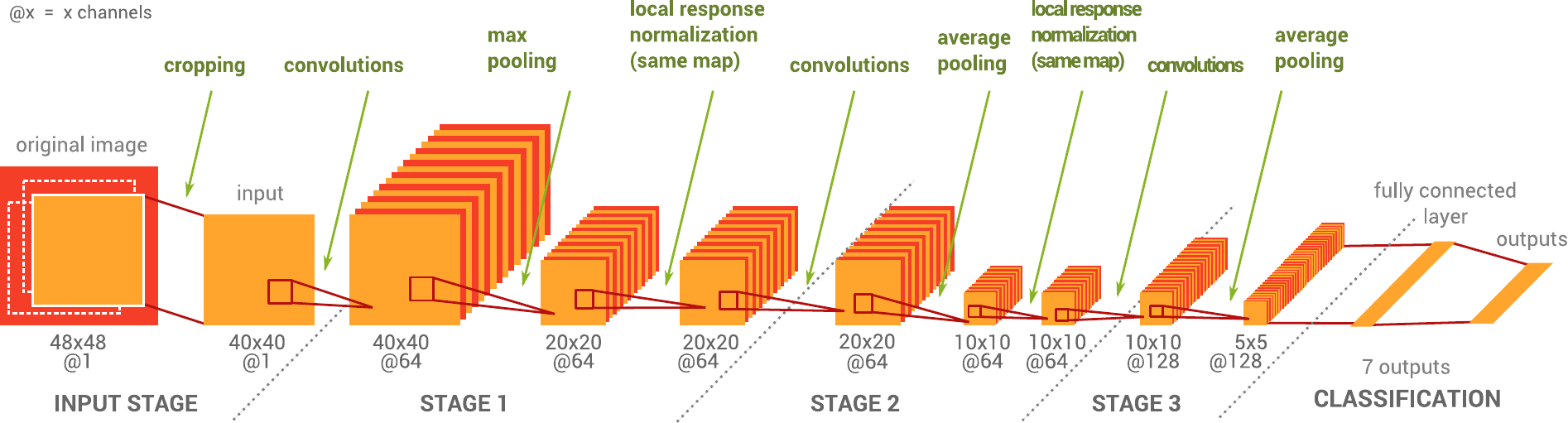}
\caption{The architecture of our ConvNet \textnumero 1.}
\label{fig:convnet_10}       
\end{figure*}

Our ConvNet uses the C++ and CUDA 
implementation written by Alex Krizhevsky \cite{Krizhevsky:2012:code} interfaced in Python. The network's 
architecture used here is presented in Figure \ref{fig:convnet_10}. The ConvNet takes batches of 
$48\times 48$ images as input and performs a random cropping into smaller 
$40 \times 40$ sub-images at each epoch. These images are then randomly flipped horizontally 
with a probability of 0.5. These two common methods allow us to expand the limited 
training set and avoid over-fitting.

The ConvNet architecture has 4 stages containing different layers. The first two
stages include a convolutional layer followed by a pooling layer, then a local 
response normalization layer \cite{NIPS2012_0534}. 
The third stage includes
only a convolutional layer followed by a pooling layer.
Max-pooling is used in the first stage, while average-pooling is used in the 
next stages. The last stage consists of seven softmax units, 
which output seven probabilities, one for each of the seven emotion labels.
The activation function used in the convolutional layers is the rectified 
linear unit (ReLU) activation function. The two first convolutional layers use 64 filters each, 
and the last one 128, all of size $5\times 5$ pixels. Each convolutional layer has the 
same learning parameters: a 0.001 learning rate for the filters and 0.002 for 
biases, 0.9 momentum for both filters and biases and a weight decay of 0.004 per
epoch. The fully-connected layer shares the same hyperparameters except for 
the weight decay, which we set to 1. These hyperparameters are the same as 
the one provided by Krizhevsky \cite{Krizhevsky:2012:code} in his example layers 
configuration files. The architecture is depicted in Figure \ref{fig:convnet_10}.

Classification at test time is done using the $40\times 40$ sub-images cropped from the 
center of the original images. We stopped learning at 453 epochs using 
early-stopping on the competition validation and train sets. As stated earlier, we only 
used extra data to train the network, and the competition training and validation 
datasets were only used for early stopping and the subsequent training of the SVM.

\ifthenelse{\boolean{showConvnet1Curvefig}}{
Figure \ref{fig:convnet1_curve} presents the learning curve for the deep 
network trained on extra data and validated on the competition training and 
validation set.
\begin{figure}
  \includegraphics[width=.98\columnwidth]{images/experiment4_curve_v3.png}
\caption{ConvNet 1 classification error on training and validation sets (before aggregation)}
\label{fig:convnet1_curve}
\end{figure}
}{}
A shallower ConvNet was explored for the 2013 competition. It performed worse 
than ConvNet 1 and we did not revisit it for the 2014 dataset. In the tables 
for the AFEW2 results, it is referred to as ConvNet 2. For details on the 
architecture see \cite{kahou2013combining}.

\subsubsection{Facetube extraction procedure}
\label{sec:models-conv1-facetube}
For the competition dataset video frames were extracted preserving the original 
aspect ratio.
Then the Google Picasa face detector \cite{googlepicasa} was used to crop 
detected faces in each frame. To get the bounding box parameters in the 
original image, we used Haar-like features for matching, because direct 
pixel-to-pixel matching did not achieve the required performance. Picasa did 
not detect faces in every frame. To fix this, we searched the spatial 
neighborhood of the temporally closest bounding box for regions with an 
approximately matching histogram of color intensities.
We used heuristics, such as the relative positioning, sizes and overlap, to 
associate bounding boxes of successive frames and generate one facetube for 
each subject in the video.

For a few clips in the competition test sets, the Picasa face detector did not 
detect any faces. So we used the combined landmark placement and face detection 
method described in \cite{ZhuRamFace} to find faces in these clips. Using the 
facial keypoints output by that model we built bounding boxes and assembled them 
into facetubes with the previously described procedure.

\paragraph{Facetube smoothing} In order to get image sequences where face sizes 
vary gradually, we applied a smoothing procedure on the competition facetube bounding 
boxes described in \ref{sec:models-conv1-facetube}. For all images of a 
facetube, coordinates of the opposite corners of the bounding boxes were 
smoothed with a 2-sided moving average (using a window size of $11$ frames). 
The largest centered squares, that fit into these smoothed bounding boxes, 
yielded new bounding boxes which more tightly frame the detected faces. To 
restrict the amount of motion of the bounding boxes the same kind of smoothing 
was also applied to the center of the bounding boxes.

Side lengths of the bounding boxes can vary due to changes of camera position 
or magnification (e.g. changing from a medium shot to a close-up shot). To be 
able to handle this, a further polynomial smoothing technique was applied 
directly on the bounding box side lengths. Two low-order polynomials of degree 
0 (constant) and 1 (linear) were fit through the side lengths of the bounding 
boxes. If the slope of the linear polynomial is above a scale threshold ({\it slope
$\cdot$ facetube length}), we use the values of the linear polynomial as side 
lengths, else we use values from the constant smoothing polynomial. Empirically,
we found that a threshold of $1.5$ yielded reasonable results.

The final facetubes were then generated by cropping based on the smoothed 
bounding boxes and resizing the patches to $48\times 48$. Per-frame emotion
label probabilities were extracted for each facetube using the ConvNet.

\subsubsection{Aggregation into video descriptors and classification}
\label{sec:models-conv1-svm}
\begin{figure}
  \centering
  \includegraphics[width=.98\columnwidth]{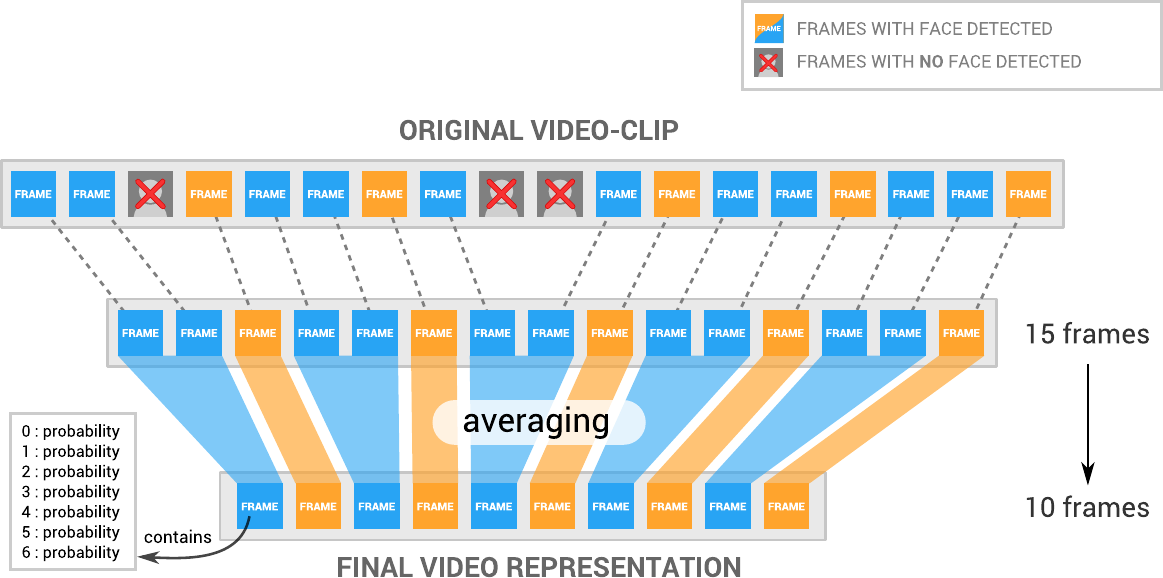}
\caption{Frame aggregation via averaging}
\label{fig:video_averaging}
\end{figure}

\begin{figure}
  \centering
  \includegraphics[width=.98\columnwidth]{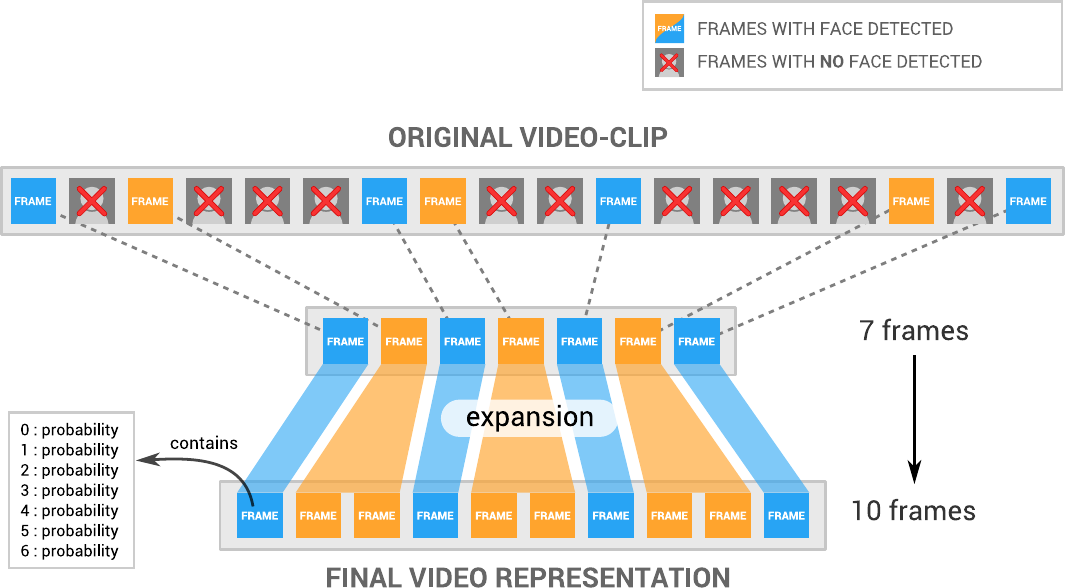}
\caption{Frame aggregation via expansion}
\label{fig:video_expansion}
\end{figure}

We aggregated the per-frame probabilities for all frames of a facetube for 
which a face was detected into a fixed-length video descriptor to be used as 
input to an SVM classifier. For this aggregation step we concatenated the 
seven-dimensional probability vectors of ten successive frames, yielding 70 
dimensional feature vectors.
Most videos have more than ten frames and some are too short and there are
frames without detected faces.
We resolved these problems using the following two aggregation approaches:
\begin{itemize}
    \item Video averaging: For videos that were too long, we averaged the 
        probability vectors of 10 independent groups of frames taken uniformly 
        along time, contracting the facetube to fit into the 10-frame video 
        descriptors. 
This is depicted in Figure \ref{fig:video_averaging}.
    \item For videos that contain too few frames with detected faces, we 
        expanded by repeating frames uniformly to get 10 frames in total. This 
        is depicted in Figure \ref{fig:video_expansion}.
\end{itemize}

\noindent The video descriptors for the training set were then used to train an
SVM (implemented by \cite{CC01a}) with a radial basis 
function (RBF) kernel. The
hyperparameters, $\gamma$ and $c$ were tuned on the competition validation 
set. The SVM type used in all experiments was a C-SVM classifier 
and the outputs are probability estimates so that the fusion with other results was simpler.

%% file: audio.tex
\subsection{Audio \& Deep Belief Networks}                                                                       
\label{sec:Audio}
As we have described earlier, deep learning based techniques have led to important successes in speech recognition \cite{hinton2012deep,graves2013speech}. In the context of emotion recognition on audio features extracted from movie clips, we used a deep learning approach for performing emotion recognition just by pretraining a deep MLP as a deep belief network (DBN) \cite{hinton2006fast}. A DBN is a probabilistic generative model where each layer can be greedily trained as a Restricted Boltzmann Machine (RBM). Initially we trained the network as a DBN in an unsupervised manner with greedy layerwise training procedure and then we used supervised finetuning\cut{ the weights by using supervised finetuning}.  In order to tune the hyperparameters of our model, we performed cross-validation using the competition validation dataset. We initially used a random search for hyperparameters and after the random search, we did manual finetuning of hyperparameters.

\subsubsection{Audio Preprocessing}
\label{sec:audio-preprocessing}
Choosing the right features is a crucial aspect of the audio classification. Mel-frequency cepstral coefficients (MFCCs) are widely used for speech recognition; however, in this task we are mainly interested in detecting emotions from the extracted audio features. 

On the other hand emotion recognition on film audio is quite different from other audio tasks. 
In addition to speech in the audio track, background noise and the soundtrack can also be significant indicators of emotion.
\cut{For example, despite the existence of speech audio in a movie, background noise and the soundtrack of the movie can be significant indicators of emotion in the clip.} For the EmotiW challenge, we extracted 29 features from each audio track using the yafee library\footnote{Yaafe: audio features extraction toolbox: \url{http://yaafe.sourceforge.net/}} with a sampling rate of 48 kHz. We used all features provided by the yafee library except ``Frames''. Additionally $3$ types of MFCC features are used, the first used $22$ cepstral coefficients, the second used a feature transformation with the temporal first-order derivative and the last one employed second-order temporal derivatives. Online PCA was applied on the extracted features, and $909$ features per timescale were retained \cite{hamel2011temporal}.

\subsubsection{DBN Pretraining} 
\label{sec:dbn-pretraining}
We used unsupervised pre-training with deep belief networks (DBN) on the extracted audio features. The DBN has three layers of RBMs, the first layer is a Gaussian RBM with noisy rectified linear unit (ReLU) nonlin\-ear\-ity \cite{dahl2013improving}, the second and third layer are both Gaussian-Bernoulli RBMs. We trained the RBMs using stochastic maximum likelihood and contrastive divergence with one Gibbs step (CD-1). 

Each RBM layer had $350$ hidden units. The first and second layer RBMs were trained with learning rates of $0.0006$, $0.0005$ and $0.001$ respectively.
An L2 penalty of $2\times10^{-3}$ and $2\times10^{-4}$ was used for the first and second layer, respectively.
Both the first and second layer RBMs were trained for 15 epochs on the competition training dataset. We bounded the noisy ReLU activations of the first layer Gaussian RBM, specifically we used the activation function: $min(\alpha, max(0, \mathbf{x}+\psi))$, where $\psi \sim N(0, \sigma(\mathbf{x}))$ with $\alpha=6$.  Otherwise large activations of the first layer RBM were causing problems training the second layer Gaussian Bernoulli RBM. We used a Gaussian model of the form $N(0, \sigma(\mathbf{x}))$, with 0 mean and standard deviation of $\sigma(\mathbf{x})=\frac{1}{1+exp(-\mathbf{x})}$. At the end of unsupervised pre-trainining, we initialized a multilayer perceptron (MLP) with the ReLU non-linearity for the first layer and sigmoid non-linearity for the second layer using the weights and biases of the DBN.

\subsubsection{Temporal Pooling for Audio Classification}
\label{sec:temporal-pooling-audio-classification}
We used a multi-time-scale learning model \cite{hamel2011temporal} for the MLP where we pooled the last hidden representation layer of an MLP so as to aggregate information across frames before a final softmax layer.
We experimented with various pooling methods including max pooling and mean pooling, but we obtained the best results with a specifically designed type of pooling for the MLP features discussed below.

Assume that we have a matrix $A$ for the activations of the MLP's last layer features that includes activations of all timescales in the clip where
$A \in R^{d_t \times d_f}$ and $d_t$ is the variable number of timescales, $d_f$ is the number of features at each timescale. We sort the columns of $A$ in decreasing order and get the top $N$ rows using the map $f:R^{d_t \times d_f} \rightarrow R^{N \times d_f}$. The most active $N$ features are summarized with a weighted average of the top-N features:
\begin{equation}
    F=\frac{1}{N}\sum_{i=0}^N w_i f^{(i)}(A;N)
\end{equation}
where $f^{(i)}(A;N)$ is the $i^{th}$ highest active feature over time and weights should be:
$\sum_{i=0}^N w_i = N$. During the supervised finetuning, we feed the reduced features to the top level softmax,
we backpropagate through this pooling function to the lower layers.
We only used the top 2 ($N=2$) most active features in the weighted average. Weights of the features were
not learned and they were chosen as $w_1=1.4, w_2=0.6$ during training and
$w_1=1.3, w_2=0.7$ during test time.
This kind of feature pooling technique worked best, if the features are extracted from a bounded nonlinearity
such as $sigmoid(.)$ or $tanh(.)$.

\subsubsection{Supervised Fine-tuning}
\label{sec:supervised-finetuning}
The competition training dataset was used for supervised fine-tuning and we applied early stopping by
measuring the error on the competition validation dataset. The features were centered prior to training.
Before initiating the supervised training, we shuffled the order of clips. During the supervised fine-tuning
phase, at each iteration on the training dataset, we randomly shuffled the order of the features in the
clip as well. At each training iteration, we randomly dropped out 98 clips from the training dataset and we
randomly dropped out 40\% of the features in the clip. 
$0.121$ \%
of the hidden units are dropped out and we used a norm constraint on the weights such that the L2 norm of the incoming
weights to a hidden unit does not exceed $1.2875$ \cite{hinton2012improving}. In addition to drop-out
and maximum norm constraint on the weights, a L2 weight penalty 
with coefficient of $10^{-5}$ was used. The rmsprop adaptive learning rate algorithm was used to
tune the learning rate with a variation of Nesterov's Momentum \cite{sutskeverimportance}. RMSProp scales down parameter updates by a running average of the gradient norm. At each iteration we keep track of the mean square of the gradients by:
\begin{equation}
RMS(\Delta_{t+1}) = \rho RMS(\Delta_{t}) + (1-\rho) \Delta_t^2
\end{equation}
and compute the momentum, then do the stochastic gradient descent (SGD) update:
\begin{equation}
    v_{t+1} = \mu v_t - \epsilon_0 \frac{\partial f(x^{(i)}; \theta_t)}{\partial \theta_t},
\qquad
\end{equation}
\begin{equation}
    \theta_{t+1} = \theta_t + \frac{\mu v_{t+1} - \epsilon_0 \frac{\partial f(x^{(i)};
                                                                            \theta_t)}{\partial
    \theta_t}}{\sqrt{RMS(\Delta_{t+1})}}
\end{equation}
After performing crossvalidation, we decided to use an $\epsilon_0=0.0005$ , $\mu=0.46$ and
$\rho=0.92$. We used early stopping based on the validation set performance, yielding an accuracy of 
$32.90$\%. Once supervised fine-tuning had completed 50 iterations, if the validation error continued
increasing, the learning rate was decreased by a factor of $0.99$.

%% file: activity.tex
\ifthenelse{\boolean{showFilters}}{
\begin{figure*}[ht]
\begin{center}
  \subfigure{\includegraphics[width=0.16\linewidth]{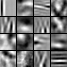}} \hspace{4mm}
  \subfigure{\includegraphics[width=0.16\linewidth]{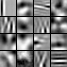}} \hspace{4mm}
  \subfigure{\includegraphics[width=0.16\linewidth]{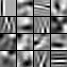}} \hspace{4mm}    
  \subfigure{\includegraphics[width=0.16\linewidth]{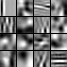}} \hspace{4mm} 
  \subfigure{\includegraphics[width=0.16\linewidth]{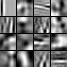}}
 \caption{Subset of filters learned by SAE model on the AFEW2 training set. Left to right: Frames 1,3,5,7 and 9.}
 \label{fig:activity_filters}
\end{center}
\end{figure*}
}{}
Given a video sequence with the task of extracting human emotion labels, it seems reasonable to also consider the temporal evolution of image frames. To this end we employ an activity recognition system for emotion recognition
based on local spatio-temporal feature computation. Using local motion features for activity recognition is a popular approach employed in many previous works \cite{ISA,KondaMM13,convGBM,Wang09evaluationof}.

Traditional motion energy models \cite{adelson1985spatiotemporal} encode spatio-temporal features of successive video frames as sums of squared quadrature Fourier or Gabor coefficients across multiple frequencies and orientations \cite{ISA}. Summing induces invariance w.r.t. content, allowing the model to yield a pure motion representation. In contrast to the motion energy view, in \cite{KondaMM13} it has been shown that the learning of transformations and introduction of invariance can be viewed as two independent aspects of learning. Based on that view, a single layered autoencoder based model named {\it synchrony autoencoder} (SAE) for learning motion representations was introduced. The classic approach is to use hand-engineered features for spatio-temporal feature extraction \cite{Wang09evaluationof}. In contrast to hand-engineered features, deep learning based methods have been shown to yield low-level motion features, which generalize well across datasets \cite{ISA,KondaMM13}. 

We use a pipeline commonly employed in works on activity recognition \cite{ISA,KondaMM13,Wang09evaluationof} with the SAE model for local motion feature computation.
We chose to use the SAE model because, compared to other learning based methods like ISA \cite{ISA} and convGBM \cite{convGBM} with complex learning rules, it can be trained very efficiently, while performing competitively.
The activity recognition pipeline follows a bag-of-words approach. It consists mainly of three modules: motion feature extraction, K-means vector quantization and a $\chi^2$ kernel SVM for classification. The SAE model acts as feature extractor. It is trained on small video blocks of size $10 \times 16 \times 16$ ($time \times rows \times columns$) randomly cropped from the competition training set. They are preprocessed using PCA for whitening and dimensionality reduction, retaining $300$ principal components. The number of randomly cropped training samples is $200,000$. The size of the SAE's hidden layer was fixed at $300$. The model was trained using SGD with a learning rate of $0.0001$ and momentum $0.9$ for $1,000$ epochs. 
\ifthenelse{\boolean{showFilters}}{The filters learned by the model on videos from the AFEW4 training set are visualized in Figure \ref{fig:activity_filters}.}{}

In past works It has been shown that spatially combining local features learned from smaller input regions leads to better representations than features learned on larger regions \cite{ISA,CoaLeeNgSingleLayer}. Here, we utilize the same method by computing local feature descriptors for sub blocks cropped from the corners of a larger $14 \times 20 \times 20$ ``super block'' and concatenating them, yielding a descriptor of motion for the region covered by the super block. PCA was applied to this representation for dimensionality reduction, retaining the first $100$ principal components. To generate descriptors for a whole video, super blocks are cropped densely for each video with a stride of $7$ on the temporal axis and $10$ on the spatial axes, i.e. with $50\%$ overlap of neighboring super blocks. 
The K-means clustering step produces a dictionary of $3000$ words, where each word represents a motion pattern. A normalized histogram over $K-means$ cluster assignment frequencies was generated for each video as input to the classifier. 

In our experiments we observed that the classifier trained on the motion features seemed to overfit on the training set and all investigated measures to avoid this problem (e.g. augmenting the data set by randomly applying affine transformations to the input videos) were also not helpful. This could be due to the videos showing little to no motion cues that correlate heavily with the emotion labels. 
The motion model by itself is not very strong at discriminating emotions, but it is useful in this task, nonetheless. It helps to disambiguate cases, where other modalities are not very confident, because it represents some characteristics of the data additional to those described by the other modalities.

%% file: bom.tex
Some emotions may be recognized from mouth features. For example, a smile often indicates happiness while an ``O''-shaped open mouth may signal surprise. For our submission, facetubes, described in section \ref{sec:models-conv1-facetube}, in resolution $96\times 96$ were cropped around a region where the mouth usually lies. This region was globally chosen by visualizing many training images, but a more precise method, such as mouth keypoint extraction \cite{ZhuRamFace}, could also be applied.

We mostly follow the method introduced by Coates et al. \cite{CoaLeeNgSingleLayer}, which achieved state-of-the-art performance on the CIFAR-10 dataset \cite{Krizhevsky09learningmultiple} in 2011, even though that method has since been superseded by convolutional networks. As a first step, each mouth image is divided into 16 equally sized sections, from which many $8\times 8$ patches are extracted. These are normalized by individually setting the mean pixel intensity to 0 and the variance to 1. After centering all patches from the same spatial region, we apply whitening, which was shown to be useful for this kind of approach \cite{CoaLeeNgSingleLayer}, keeping 90\% of the variance. For each of the 16 regions, 400 centroids are found by applying the k-means algorithm on the whitened patches.

For any given image, patches are densely extracted from each of the 16 regions and pre-processed as described above. Each patch is assigned a 400-dimensional vector by comparing it to the centroids with the triangle activation function \cite{CoaLeeNgSingleLayer}, where the Euclidean distance $z_k$ between the patch and each centroid is computed, as well as the mean $\mu$ of these distances. The activation of each feature is given by $max(0, \mu-z_k)$, so that only centroids closer than the mean distance are assigned a positive value, while distant ones stay at 0. As we have a 400-dimensional representation for each patch, the image representation would become extremely large if we simply concatenated all feature vectors. For this reason, we pool over all features of a region to get a local region descriptor. The region descriptors are then concatenated to obtain a 6,400 dimensional representation of the image.

This pooling generally uses the average activation of each feature, although we also tried taking the standard deviation across patches for each feature.
A regularized logistic regression classifier is trained on a frame-by-frame basis with the pooled features as input. When classifying a test video, the predictions of the model are averaged over all its frames.

%% file: combination.tex
\begin{table*}
\caption{Our 7 submissions with training, validation and test accuracies for the EmotiW 2013 competition.}
\begin{center}
    \begin{tabular}{rllll}
    \hline\noalign{\smallskip}
Sub. & Train & Valid & Test &  Method \\ \hline
1 & 45.79 & 38.13 & 35.58 & Google data \& TFD used to train ConvNet 1, SVM trained on aggregated frame scores \\ 
2 & 71.84 & 42.17 & 38.46 & ConvNet 1 (from submission 1) combined with Audio model using another SVM \\ 
3 &  97.11 & 40.15 & 37.17 & Mean prediction from: Activity, Audio, Bag of mouth, ConvNet 1, ConvNet 2 \\ 
4 &  98.68 & 43.69 & 32.69 & SVM with detailed hyperparameter search: Activity, Audio, Bag of mouth, ConvNet 1 \\
5 &  94.74 & 47.98 & 39.42 & Short uniform random search: Activity, Audio, Bag of mouth, CN1, CN1 + Audio \\ 
6 &  94.74 & 48.48 & 40.06 & Short local random search: Activity, Audio, Bag of mouth, CN1, CN1 + Audio \\ 
7 &  92.37 & {\bf 49.49} & {\bf 41.03} & Moderate local random search: Activity, Audio, Bag of mouth, CN1, CN1 + Audio \\ 
\hline\noalign{\smallskip}
    \end{tabular}
    \end{center}
\vspace{-5pt}
\label{table:submissions2013}
\end{table*}

\begin{table*}
\caption{Our selected submissions with test accuracies for the EmotiW 2014 competition.}
\begin{center}
    \begin{tabular}{rll}
    \hline\noalign{\smallskip}
Sub. & Test &  Method \\ \hline
 1 & 39.80 & Trained model on 2013 data, BoM failed due to different data format and replaced by uniform  \\ 
 2 & 37.84 & Trained model on 2013 data, re-learning random search without failed BofM \\ 
 3 & 44.71 & ConvNet 1 + Audio model combined with SVM, all trained on train+valid \\ 
 4 & 41.52 & ConvNet 1 + Audio model combined with SVM trained on swapped predictions \\ 
 5 & 37.35 & Google data \& TFD used to train ConvNet 1, frame scores aggregated with SVM \\
 6 & 42.26 & All models combined with SVM trained on validation predictions \\ 
 7 & 44.72 & All models combined with random search optimized on validation predictions\\
 8 & 42.51 & Only two models were trained on train+validation in combination, others used train set only\\
 9 & {\bf 47.67} & All models combined with random search optimized on full swapped predictions\\
10 & 45.45 & Bagging of 350 models similar to submission 9\\
\hline\noalign{\smallskip}
    \end{tabular}
    \end{center}
\vspace{-5pt}
\label{table:submissions2014}
\end{table*}

In figure \ref{fig:model-conf} (a-d) we show the validation set confusion 
matrices from the models yielding the highest AFEW4 validation set accuracy for each of the techniques discussed in section \ref{sec:models}. A second convolutional network for faces (Convnet \#2), which we explored, is not presented here as it obtained lower performance compared to Convnet \#1 and used similar information to make its predictions. A more detailed analysis of Convnet \#2 and comparisons on AFEW2 can be found in \cite{kahou2013combining}, but we provide some highlights here.
\paragraph{AFEW2}
From our experiments with AFEW2 we found that ConvNet1 yielded the highest validation set accuracy. We therefore selected this model as our first submission and it yielded a test set accuracy of {\bf 35.58}\%. This is also indicated in table \ref{table:submissions2013} which contains a summary of all our submissions.
ConvNet2 was our second highest performer, followed closely by the bag of mouth and audio models at 30.81\%, 30.05\% and 29.29\% respectively. 

\paragraph{AFEW4}
Here again our ConvNet1 model achieved the best results on the validation set for AFEW4. It was followed by our audio model which here yields higher performance than the bag of mouths model by a good margin, at 34.20\% and 27.42\% accuracy respectively.

We explored the strategies outlined in Sections \ref{sec:combine-avg}, \ref{sec:combine-svmmlp} and \ref{sec:combine-random} to combine models for the AFEW2 evaluation. Section \ref{sec:combine-trainonvalid} presents the strategy we used for our experiments with the AFEW4.

\subsection{Averaged Predictions -- AFEW2}
\label{sec:combine-avg}
A simple way to make a final prediction using several models is to take the average
of their predictions. We had 5 models in total, which gives 
$\sum_{i=1}^n\binom{n}{i}=31$ possible combinations (order has no importance). 
In this context it is possible to test all combinations on the validation set to find those
which are the most promising. 

\input{confusion_figure}

Through this analysis we found that the average of all models yielded the highest validation set performance of 40.15\% on AFEW2.
The validation set confusion matrix for this model is shown in figure \ref{fig:model-conf-combi} (a).
For our third 2013 submission we therefore submitted the results of the averaged predictions of all models, yielding {\bf 37.17}\% on the test. From this analysis we also found that the exact same validation set performance was also obtained with an
average not including our second convolutional network, leading us to make the conclusion
that both convolutional networks were providing similar information. We thus left it out for subsequent strategies and experiments on the AFEW4.

The next highest performing simple average was 39.90\% and consisted of simply combining ConvNet 1 and our audio model. Given this observation and the fact that the conference baselines included both video, audio and combined audio-video models we decided to submit a model in which we used only these two models. However, we first explored a more sophisticated way to perform this combination. 

\subsection{SVM and MLP Aggregation Techniques -- AFEW2}
\label{sec:combine-svmmlp}
To further boost the performance of our combined audio and video model we simply 
concatenated the results of our ConvNet 1 and audio model using vectors and learned a SVM 
with an RBF kernel using the challenge training set. The hyperparameters of the
SVM were set via a two stage coarse, then fine grid search over integer powers of 10, then
non-integer powers of 2 within the reduced region of space. The hyperparameters correspond 
to a kernel width term, $\gamma$ and the $c$ parameter of SVMs. This process
yielded an accuracy of 42.17\% on the validation set, which became our second submission
and produced a test accuracy of {\bf 38.46}\%. 

Given the success of our SVM combination strategy, we tried the same technique using the predictions of all models.
However, this process quickly overfit the training data and we were not able to produce any models that 
improved upon our best validation set accuracy obtained via the ConvNet 1 and audio model. We observed a similar effect
using a strategy based upon an MLP to combine the results of all model predictions.

We therefore tried a more sophisticated SVM hyperparameter search to re-weight
different models and their predictions for different emotions. We implemented this
via a search over discretized $[0,1,2, 3]$ per dimension scaling factors. While this resulted
in 28 additional hyperparameters this discretization strategy allowed us to explore all combinations.
This more detailed hyperparameter tuning did allow us to increase the validation set performance
to 43.69\%. This became our fourth 2013 submission; however, the strategy yielded a decreased
test set performance at {\bf 32.69}\%.

\subsection{Random Search for Weighting Models -- AFEW2}
\label{sec:combine-random}
Recent work \cite{bergstra2012random} has shown that random search for hyperparameter optimization
can be an effective strategy, even when the dimensionality of
hyperparameters is moderate (ex. 35 dimensions).
Analysis of our validation set confusion matrices shows that different models have very different performance characteristics across the different emotion types. We therefore formulated the re-weighting of per-model and per-emotion predictions as a hyperparameter search over simplexes, weighting the model predictions for each emotion type.

\begin{figure}[h!]
  \centering
  \subfigure[{\small Emotiw 2013}]{
    \includegraphics[width=\columnwidth]{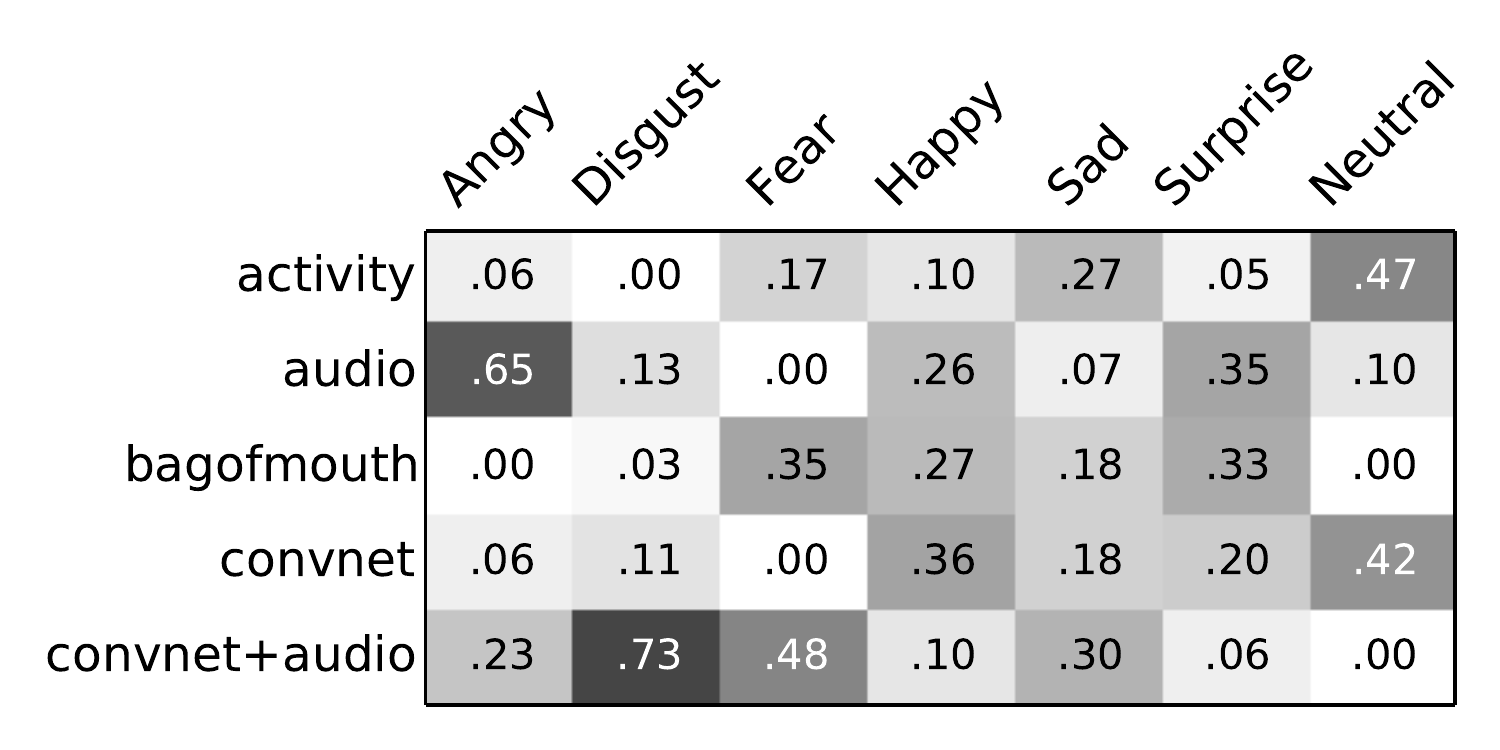}
  }
  \subfigure[{\small EmotiW 2014}]{
    \includegraphics[width=\columnwidth]{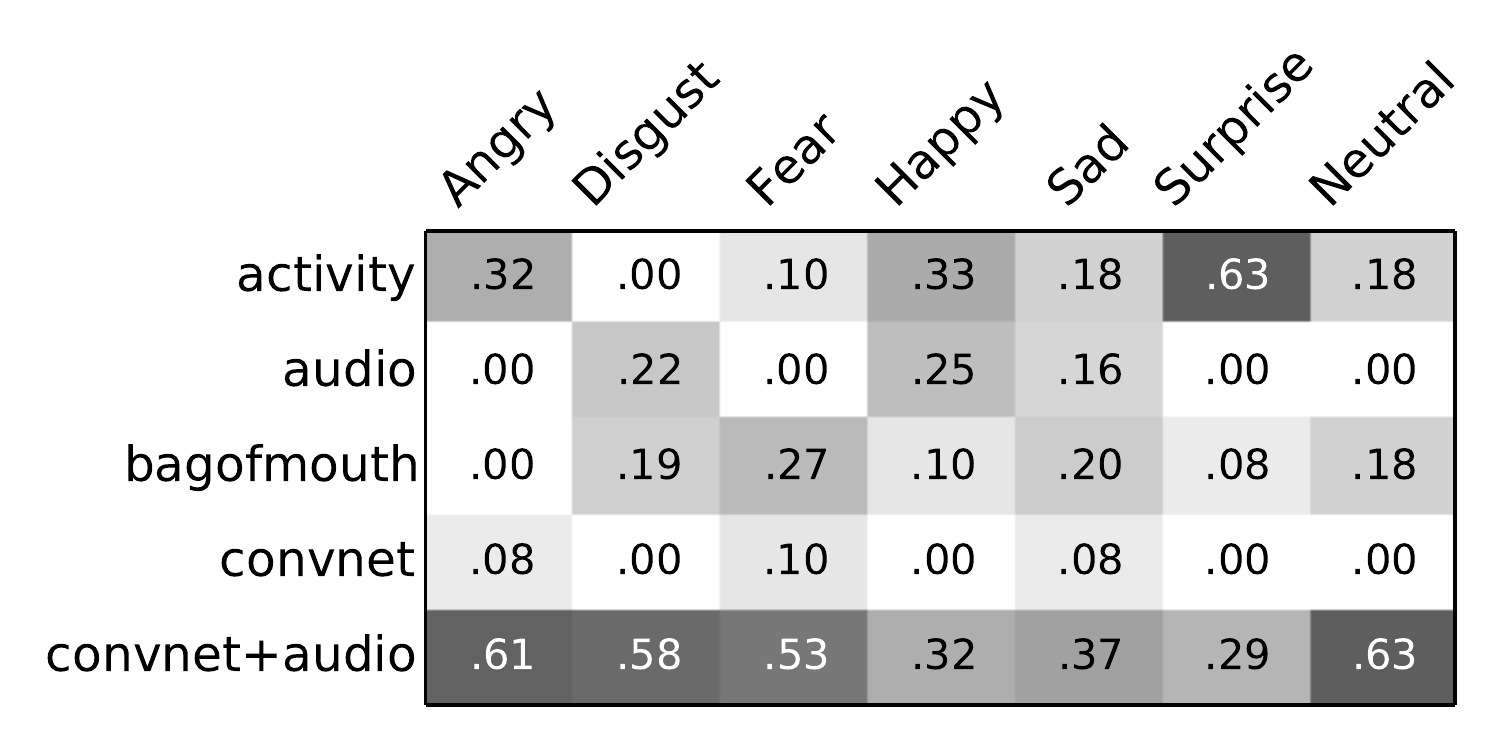}
  }
\caption{Final weights used for model averaging in our best submissions.}
\label{fig:finalw}
\end{figure}

To perform the random search, we first sampled random weights from a uniform distribution and then normalized them
to produce seven simplexes. This process is slightly biased towards weights that are less extreme compared to 
other well known procedures that are capable of generating uniform values on simplexes.
After running this sampling procedure for a number of hours we used the weighting that yielded the highest 
validation set performance (47.98\%) as our 5th 2013 submission. This yielded a test set accuracy of {\bf 39.42}\%.
We used the results of this initial random search to initiate a second, local search procedure which is analogous in a sense
to the typical two level coarse, then fine level grid search used for SVMs. In this procedure we generated random weights 
using a Gaussian distribution around the best weights found so far. The weights were tested by calculating 
the accuracy of the so-weighted average predictions on the validation set.
We also rounded these random weights to 2 decimals to help to avoid overfitting on the validation set. 
This strategy yielded {\bf 40.06}\% test set accuracy with a short duration search and {\bf 41.03}\% with a longer search - our best performing 2013 submission on the test. 
The validation set confusion matrix for this model is shown in figure \ref{fig:model-conf-combi} (b) and the weights obtained through this process are shown in figure \ref{fig:finalw} (a).
\vspace{-5pt}

\subsection{Strategies for the Emotiw 2014 Challenge and the AFEW4 Data}
\label{sec:combine-trainonvalid}
While we did not participate in the EmotiW 2014 challenge we have performed a sequence of experiments using the underlying AFEW4 dataset and the training, validation and test sets partitions defined by the challenge organizers. We have performed these experiments after the challenge period so as to explore the behaviour of our general technique as well as some different training strategies arising from the fact that the challenge is defined differently.
Specifically, in the EmotiW 2014 challenge it is permitted to re-train all models using the combined training and validation set if desired.  We correspondingly explored the following set of strategies.  

As an initial step, we simply re-ran our best model from the 2013 competition, without retraining it on the 2014 competition dataset. Predictions of Bag-of-mouth model were replaced by uniform distribution. Our Bag-of-mouth model was trained on faces provided by the organizers which were RGB in 2013 and grayscale in 2014, this caused the model to fail on new dataset.
Using models trained on AFEW2, we computed predictions on AFEW4 test set, which gave 39.80\% accuracy. The 1\% loss could possibly be attributed to the substitution of the Bag-of-mouth model with uniform distribution. However, sound comparison with previous results cannot be made as AFEW2 and AFEW4 test sets are different. Retraining the combination model on all models trained on AFEW2 but bag-of-mouth resulted in a lower 37.84\% accuracy. We used a more aggressive random search procedure by starting hundreds of random searches with different initializations. The generalization decrease from submission 1 to 2 was most likely caused by overfitting because of this aggressive random search. Nevertheless, as AFEW4 training and validation sets are larger than their AFEW2 relatives, models trained on the latter might not be competitive in the Emotiw 2014 Challenge. Therefore, we trained our models on AFEW4 data for submission 3 to 10.

In preparation for the following sets of experiments
all sub-models were trained on training set and validation set alone. They were also trained on training set combined with validation set. This yields three different sets of predictions from which one may explore and compare different training and combination strategies. Training on the training set and the validation set separately allowed us to easily do 2-fold cross-validation, while training on all data combined is a commonly used strategy to exploit all available data but can involve different techniques for setting model hyperparameters.

\subsubsection{An SVM combination approach using all data}
One simple method for learning when working with a single training set and a single validation set is to use the training set to train a model multiple times with different hyperparameters, then select the best model using the validation set. One can then simply use these hyperparameter settings and retrain the model using the combined training and validation set.
This method is known to work well in practice.

We first used this method to train an SVM to combine the predictions of the ConvNet1 model and the audio model. It resulted in  44.71\% test accuracy, an impressive 7\% improvement over ConvNet1 alone (37.35\%) and 6\% improvement over the same combination trained only on the 2013 AFEW2 training set (38.26\%). 
An important factor might be that we are using predictions on data not seen during sub-model training to train the combination model. That is, they are less biased than training predictions, which makes it possible for the SVM to generalize better. The validation set alone is, however, too small to train a good combination model.

To capitalize on this effect, we trained another SVM on swapped predictions, i.e. the predictions on the validation set came from sub-models trained on training set and predictions on training set came from sub-models trained on the validation set. An SVM was trained on both swapped sets separately to select the best hyper-parameters before training a final SVM on all swapped predictions. With 41.52\% test accuracy, this model is worse then the previous one (44.71\%). A possible reason for this is that the training and validation sets are unbalanced and relatively small. 
Good sub-models trained on the larger training set tend to generate good predictions on small validation sets, while worse sub-models trained on the small validation set generate worse predictions on the bigger training set. An obvious solution would be to generate swapped predictions in a manner similar to leave-one-out cross-validation, the drawback is that for our setting we would need to train 5 times ~900 models on each fold to generate the predictions for the meta-model.

Finally, similar to section \ref{sec:combine-random}, we trained the SVM only on validation data. We hoped training an SVM would yield results similar to running random search. It did not. As explained in next section, running random search on the validation set predictions gives 44.72\% while training an SVM on same data gives only 42.26\%.

\subsubsection{Weighting models and random search using all data}

A random search procedure for determining the parameters of a linear per-class and per-model weighting was computed as described in section \ref{sec:combine-random}, but for the AFEW4 (EmotiW 2014 challenge data). 
For our first experiment we run a random search using the validation set predictions, then used the resulting weights to compute the weighted average of predictions of sub-models trained on all data. To be clear, the only difference to our best model from 2013 submissions, was that we applied the weighted average on sub-models trained on the combined training and validation set of the 2014 dataset. This yielded a test accuracy of 44.72\%, 2\% higher than the same procedure with SVM training, but no gain over the best combination of ConvNet1 with audio models (44.71\%).

Random search can also be applied to swapped predictions such as those explained in the previous section. Running random search on such predictions gave our best results on AFEW4, 47.67\%, slightly higher than the first runner up in the EmotiW 2014 competition \cite{sun2014combining}. The weights found through this procedure are shown in Figure \ref{fig:finalw} (b).
A comparison of test accuracies for both the 2013 and 2014 EmotiW datasets with other methods is shown in table \ref{tab:afew2}.

As some models were overfitting to the training data, we tried to separate overfitters from the other models and combine them together. We ran a random search on ConvNet1, Bag-of-mouth and activity recognition predictions of validation data. Then we ran a second random search on top of their weighted average with our ConvNet1+Audio SVM combination of submission 3. This final weighted average was used to compute the test predictions, giving only 42.51\%. 

Weights found by random search varied a lot from one run to another. We tried bagging of 350 independent weighted averages found by random searches similar to submission 9 (which obtained 47.67\%). Surprisingly, the bagging approach achieved a lower accuracy of 45.45\%, our second best result on AFEW4.

%% file: confusion_figure.tex
\begin{figure}[!hbp]
  \centering
  \subfigure[{\small ConvNet 1, (45.97, 42.33, 37.35*)}]{
    \includegraphics[width=.46\columnwidth]{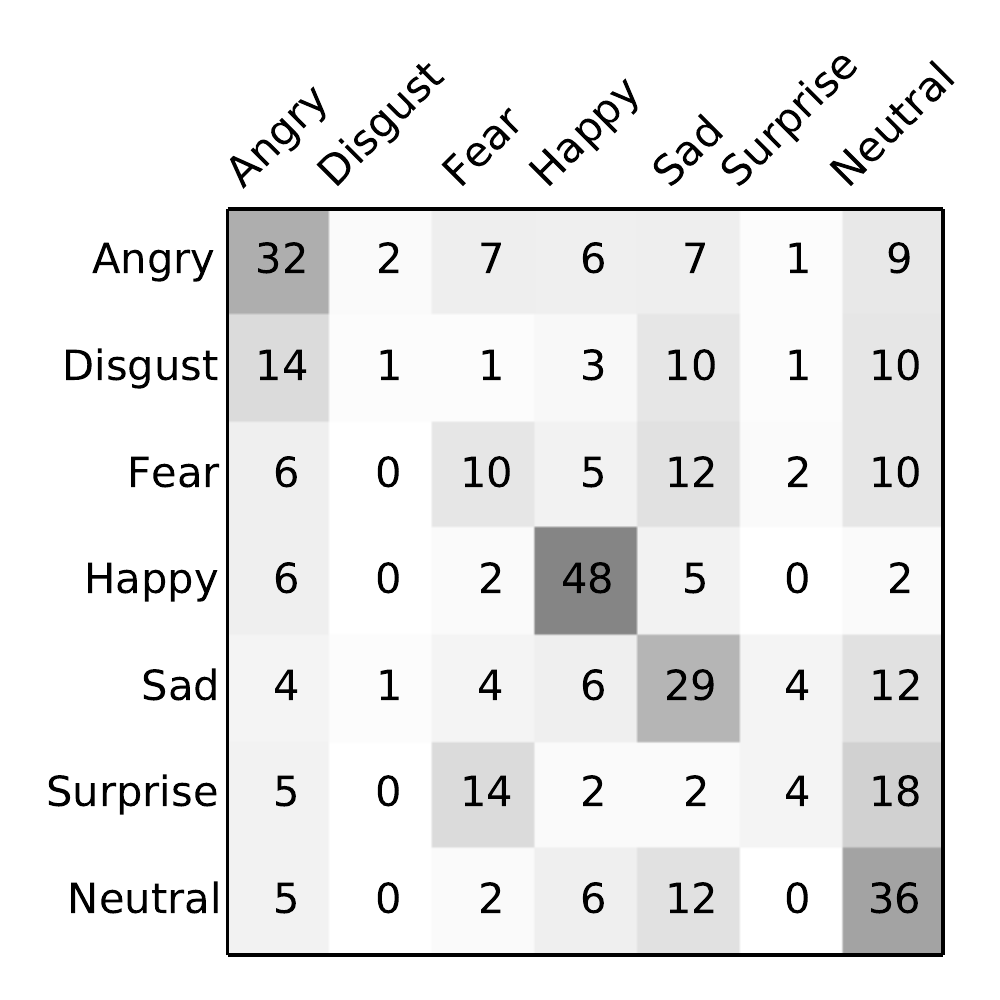}
  }
\vspace{-.15in}
  \subfigure[{\small Audio, (53.46, 34.20,-)}]{
    \includegraphics[width=.46\columnwidth]{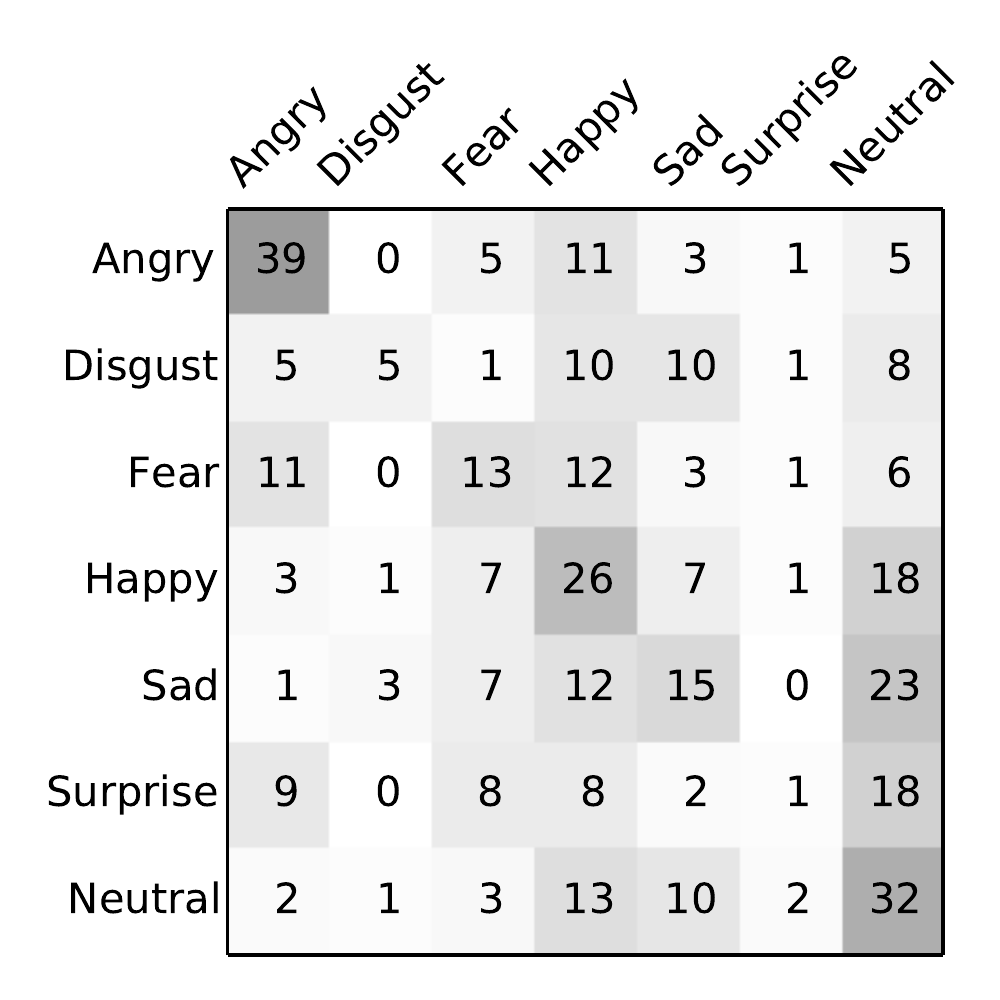}
  }
  \subfigure[{\small Activity Rec., (46.37, 25.07,-)}]{
    \includegraphics[width=.46\columnwidth]{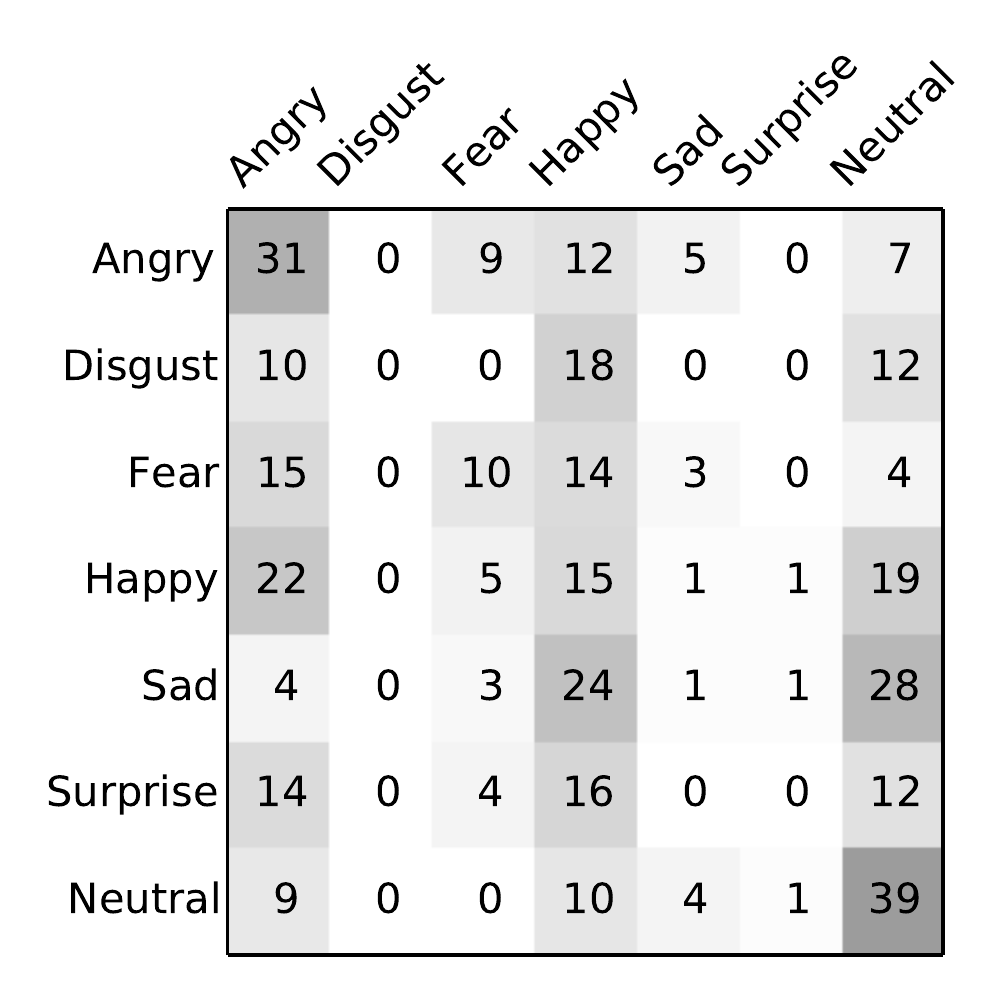}
  }
\vspace{-.15in}
  \subfigure[{\small Bag of mouth, (93.08, 27.42,-)}]{
    \includegraphics[width=.46\columnwidth]{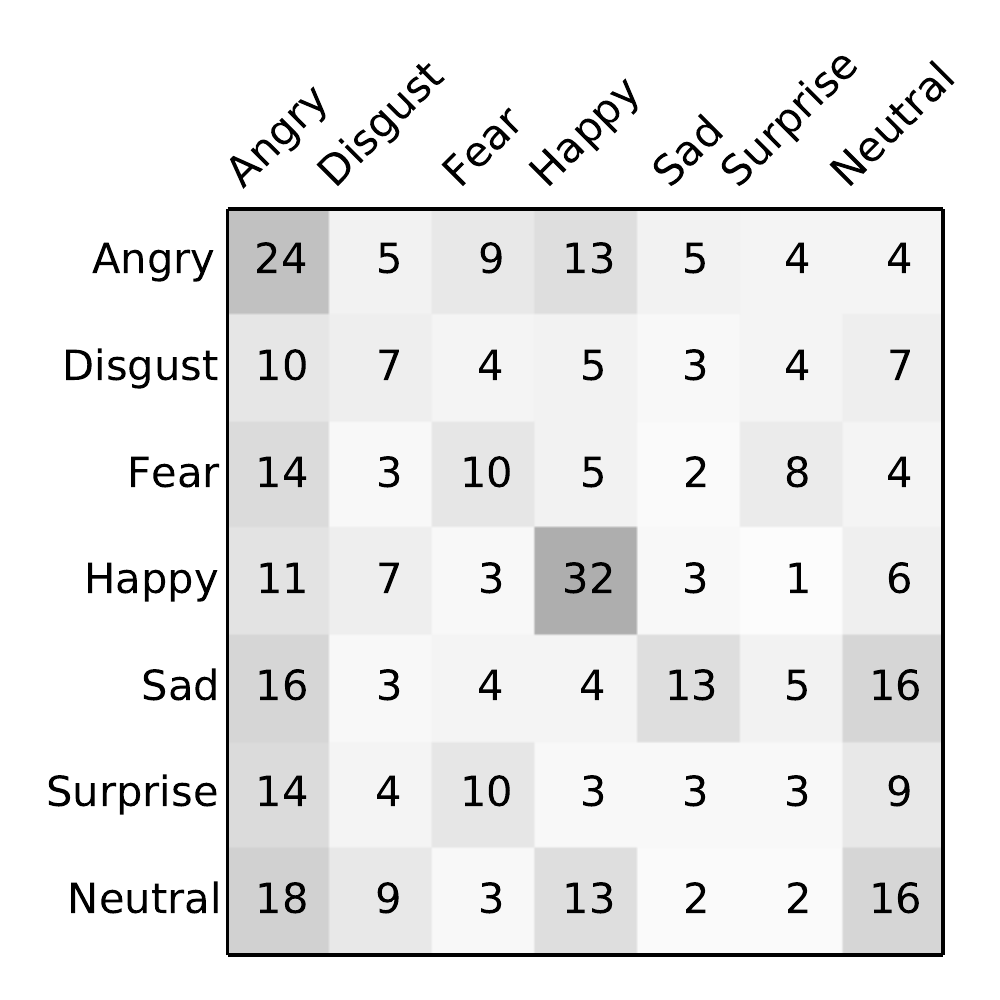}
  }
  \caption{
  Confusion matrices for the AFEW4 validation set. 
  Accuracies for each method are specified in parentheses
  (training, validation \& test sets, if applicable).
  *Model has been retrained on both training and validation set prior to testing
  \vspace{1cm}}
  \label{fig:model-conf}
\end{figure}
\begin{figure}[!hbp]
  \centering
    \subfigure[{\small Simple Average of Models, (97.11, 40.15, 37.17)}]{
    \includegraphics[width=.46\columnwidth]{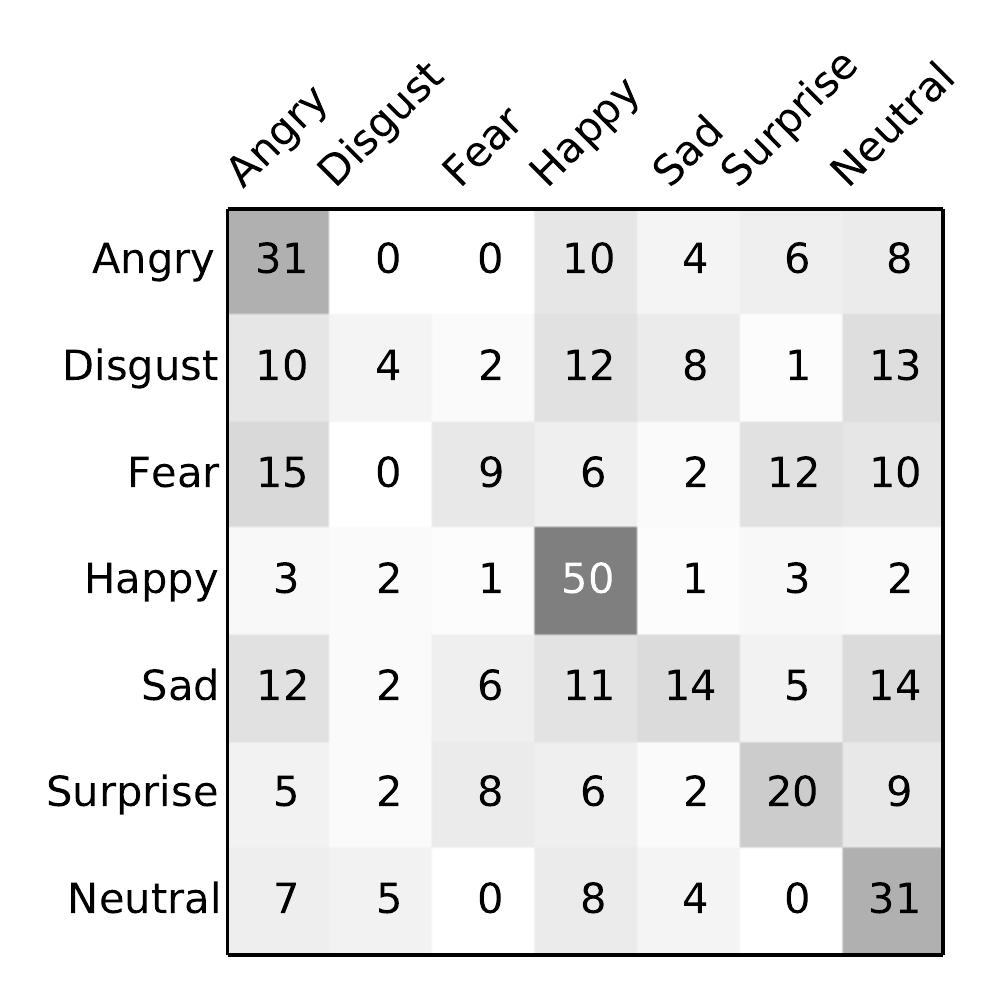}
  }
\vspace{-.15in}
  \subfigure[{\small Random Search on Weighted Avg.,       (92.37, 49.49, 41.03)}]{
    \includegraphics[width=.46\columnwidth]{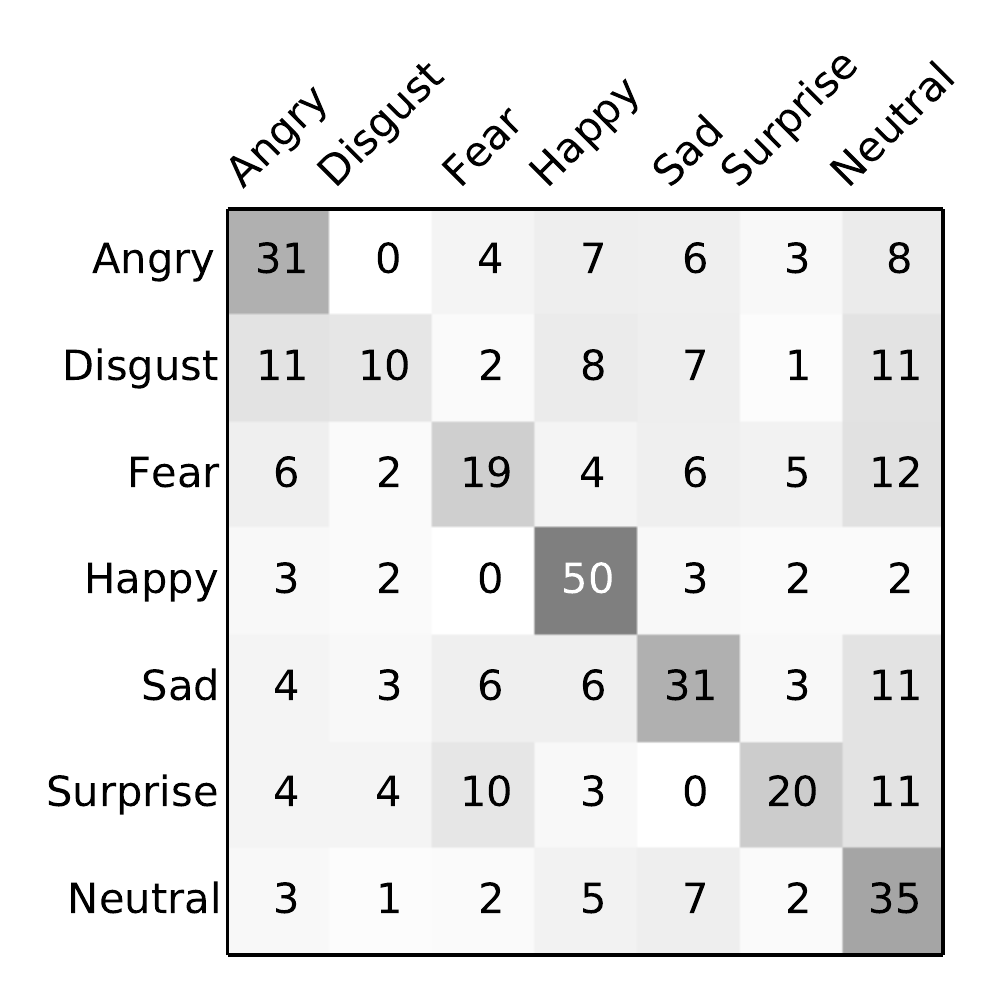}
  }
  \subfigure[{\small ConvNet 1 \& Audio, \hspace{.2in}\ (-, -, 44.71*)}]{
    \includegraphics[width=.46\columnwidth]{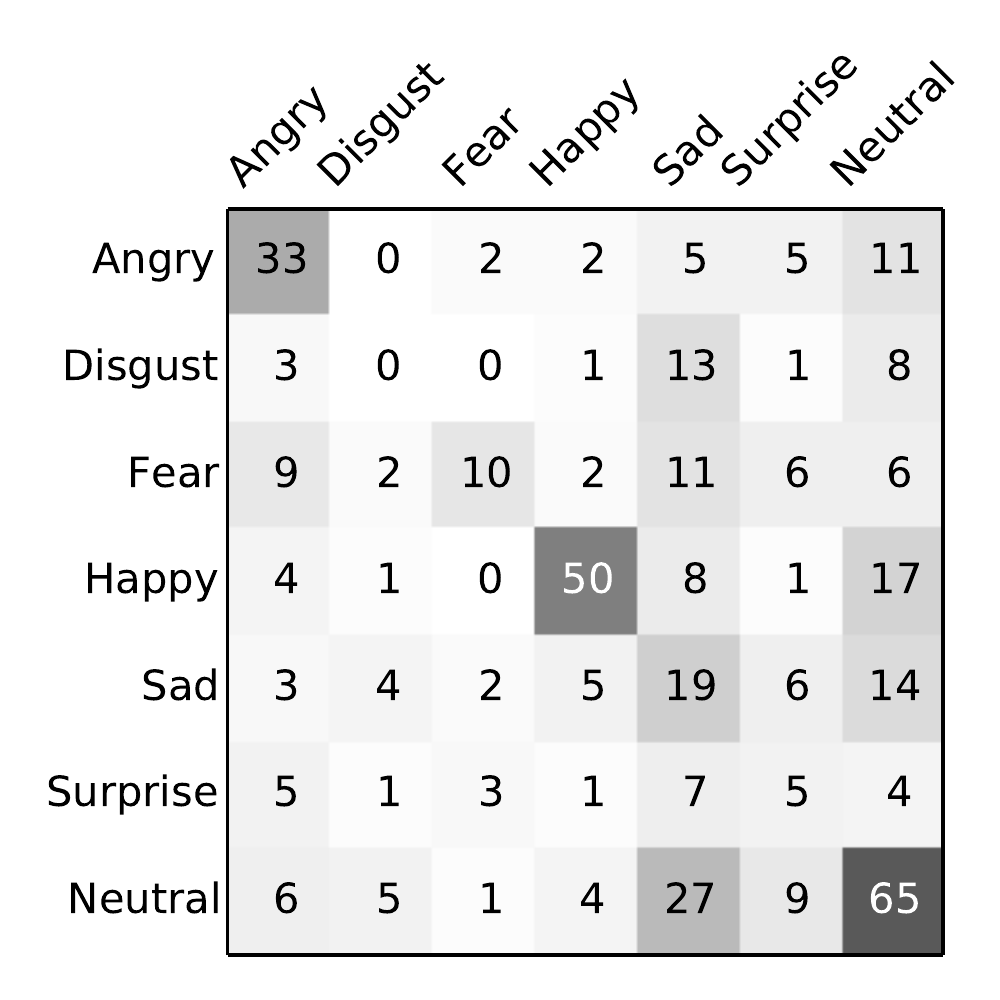}
  }
\vspace{-.15in}
  \subfigure[{\small All modalities  (submission 9), \cut{\hspace{.2in}}\ (-, -, 47.67*)}]{
    \includegraphics[width=.46\columnwidth]{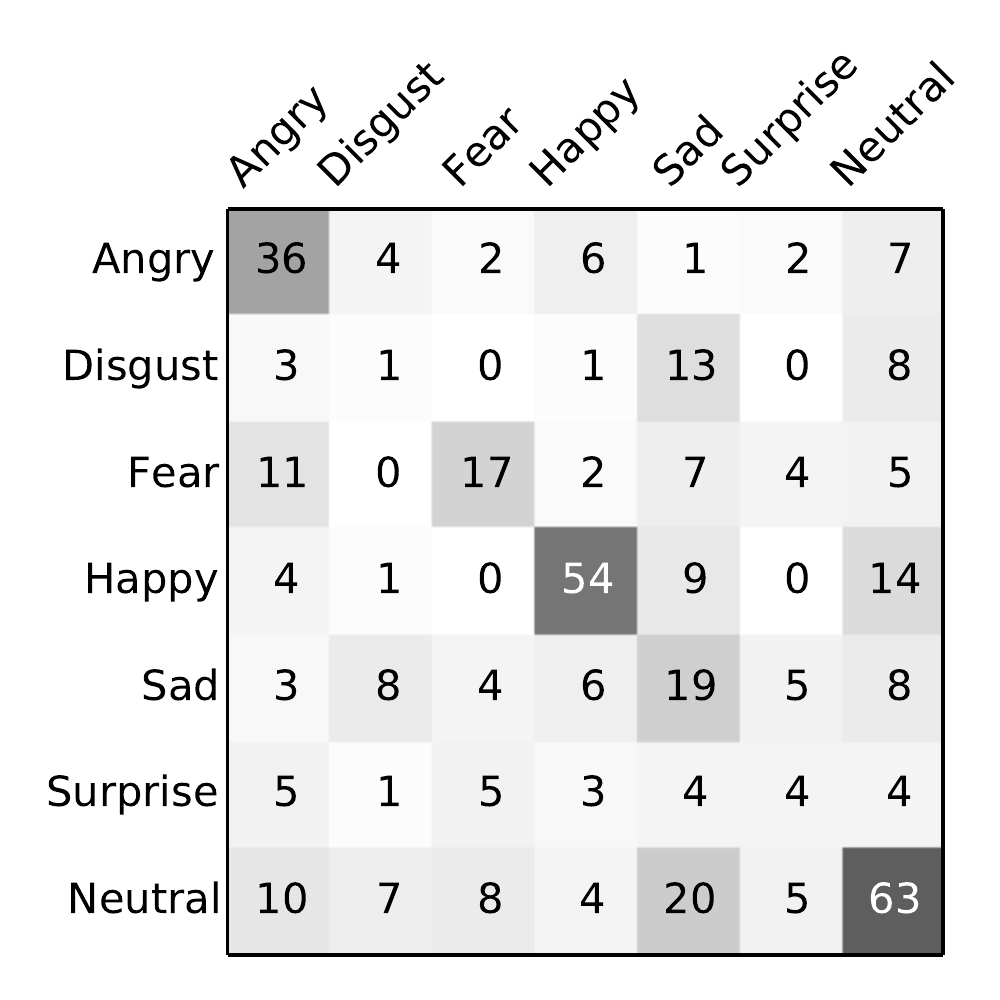}
  }
  \caption{
  Confusion matrices on the test set of AFEW2 (a-b) and AFEW4 (c-d). 
  Accuracies for each method are specified in parentheses
  (training, validation \& test sets, if applicable).
  *Model has been retained on both training and validation set prior to testing
  \vspace{1cm}}
  \label{fig:model-conf-combi}
\end{figure}

%% file: conclusion.tex
Our experiments with both competition datasets (2013 and 2014) have lead to a number of contributions and insights which we believe may be more broadly applicable. 
First, we believe that our approach of using the large scale mining of imagery from Google image search to train our deep neural network has helped us to avoid overfitting to the provided challenge dataset.

We achieved better performance when we used the competition data exclusively for training the classifier and used additional face image data for training of the convolutional network. The validation set accuracy was significantly higher than in our experiment in which we trained the network directly on extracted faces from the challenge data.
It is our intuition that video frames in isolation are not always representative of the emotional tag assigned to the clip, and using one label for video length introduces noise to the training set. 
In contrast, our additional data contained only still images with a clear correspondence between image and label.
The problem of overfitting had both direct consequences on per-model performance on the validation set as well as indirect consequences on our ability to combine model predictions. Our analysis of simple model averaging showed that no combination of models could yield superior performance to an SVM applied to the outputs of our audio-video models. Our efforts to create both SVM and MLP aggregation models lead to similar observations in that models quickly overfit the training data and no settings of hyperparameters could be found which would yield increased validation set performance. We believe this is due to the fact that the activity recognition and bag of mouth models severely overfit the challenge training set and the SVM and MLP aggregation techniques - being quite flexible - overfit the data in such a way that no traditional hyperparameter tuning could yield validation set performance gains.

These observations led us to develop the novel technique of aggregating the per model and per class predictions via random search over simple weighted averages.  
The resulting aggregation technique is therefore of extremely low complexity and the underlying prediction was therefore highly constrained - using simple weighted combinations of complex deep network models, each of which did reasonably well at this task. We were therefore able to explore many configurations in a space of moderate dimensionality quite rapidly as we did not need to re-evaluate the predictions from the neural networks and we did not adapt their parameters. 
As this obtained a marked increase in performance on both the challenge validation and test sets, it lead us to the following interpretation: Given the presence of models that overfit the training data, it may be better practice to search a moderate space of simple combination models. This is in contrast to traditional approaches such as searching over the smaller space of SVM hyperparameters or even a moderately sized space of traditional MLP hyperparameters including the number of hidden layers and the number of units per layer.